# Temporal speckle enables probabilistic weights for uncertainty-aware photonic AI

F. Brückerhoff-Plückelmann[1*], H. Borras[2*], S. U. Hulyal[3*], L. Meyer[1], X. Ji[3], J. Hu[3], J. Sun[3], B. Klein[2], F. Ebert[1], J. Dijkstra[1], L. McRae[1], P. Schmidt[1], T. J. Kippenberg[3†], H. Fröning[2], W. Pernice [1†]

[1]*Kirchhoff-Institute for Physics, University of Heidelberg; Heidelberg, 69120, Germany.*

[2]*Institute of Computer Engineering, University of Heidelberg; Heidelberg, 69120, Germany.*

[3]*Institute of Physics, Swiss Federal Institute of Technology, Lausanne, CH-1015 Lausanne, Switzerland.*

[*]These authors contributed equally to this work.

[†]Corresponding author. Email: wolfram.pernice@kip.uni-heidelberg.de, tobias.kippenberg@epfl.ch

**Artificial intelligence increasingly guides decisions in safety-critical settings, yet neural networks can remain overconfident when confronted with ambiguous observations or unfamiliar inputs. Bayesian neural networks quantify such uncertainty by replacing every weight with a probability distribution, but sampling many distributions at prediction time overwhelms deterministic digital hardware. Here we show that the temporal speckle of broadband light provides a native physical representation of probabilistic weights. Random beating among the frequency components of individually amplitude-bandwidth programmed spectral channels enables optical-encoding of mean and variance in each channel independently. This way, a single broadband light source stores many probabilistic weights in parallel and delivers mutually uncorrelated samples without requiring computationally demanding random-number generation. We harness this mechanism in an ultrafast photonic processor that computes one probabilistic convolution every 37.5 ps and allows simultaneous classification and out-of-distribution detection in hybrid Bayesian neural networks. Our system classifies blood-cell images while recognizing withheld cell types as unfamiliar, and separates epistemic novelty from aleatoric ambiguity on standard uncertainty benchmarks. Programming probability distributions into the fluctuations of light removes the sampling bottleneck of Bayesian inference and opens a route to high-speed, uncertainty-aware photonic AI.**

## Introduction

Artificial intelligence (AI) increasingly informs decisions in medicine, autonomous systems and other safety-critical settings, where an incorrect prediction can have serious consequences. Reliable deployment therefore requires models that recognize uncertainty arising from noisy or ambiguous observations (aleatoric uncertainty) and from inputs that differ from the training data (epistemic uncertainty) as illustrated in Fig. 1a. Conventional deterministic neural networks can remain confident in both situations and may return an incorrect prediction without warning [1, 2]. Bayesian neural networks address this limitation by inferring probability distributions over model parameters from the training data rather than single point estimates (Fig. 1b) [3, 4, 5, 6]. During prediction, repeated stochastic forward passes sample these distributions and produce a predictive distribution from which uncertainty can be quantified. This principled treatment of uncertainty, however, comes with substantial computational overhead from repeated sampling and probabilistic operations, limiting the use of Bayesian neural networks in complex, real-time applications. Addressing these limitations would not only accelerate AI, but also pave the way for a new generation of intelligent systems capable of reasoning under uncertainty instead of ignoring it.

The corresponding hardware challenge is not merely to generate random numbers, but to represent many independently programmable probability distributions and process their samples efficiently. Conventional digital accelerators generate and transfer stochastic weights between memory and compute units, increasing latency, energy consumption and memory traffic [7]. Analogue hardware can instead represent probabilistic parameters through intrinsic physical fluctuations and compute with them in situ. Such systems require stochastic signals whose statistics are independently tunable, stable and scalable across many weights. Electronic implementations based on memristive devices can control the mean and variance of stochastic conductances, but often require multiple devices per weight and remain affected by device variability, drift and limited bandwidth [8, 9, 10]. Photonics offers substantially larger bandwidth and sub-nanosecond processing [11], and previous photonic Bayesian approaches have explored programmed optical noise, chaotic light and interferometric stochastic devices [12, 13, 14]. What has remained missing is a physical mechanism that enables programmable weight distributions, parallel sampling and direct photonic computation while remaining compatible with established machine-learning frameworks.

Here we show that spectrally shaping broadband chaotic light provides such a mechanism. We encode each probabilistic weight in an individual frequency channel, using the integrated optical power to program its mean and the channel bandwidth to program its standard deviation (Fig. 1c). Combining this spectral encoding with high-speed temporal modulation and the frequency-dependent delay of an integrated chirped waveguide Bragg grating enables a photonic probabilistic convolution accelerator that computes one convolution output every 37.5 ps. A PyTorch-compatible interface connects the accelerator to deterministic digital layers and supports hardware-aware training of hybrid Bayesian neural networks. We experimentally deploy the resulting system for uncertainty-aware blood-cell classification, detecting erythroblasts withheld from training as epistemically uncertain. We further evaluate the hybrid photonic network on MNIST, Fashion-MNIST and Ambiguous-MNIST, demonstrating the separation of epistemic novelty from aleatoric ambiguity [15]. These results establish spectrally encoded probabilistic weights as a route to high-speed uncertainty-aware photonic artificial intelligence.

## Spectral encoding of probabilistic weights

We encode each probabilistic neural-network weight in a separate frequency channel of broadband chaotic light (Fig. 1b,c). The integrated channel power $P_i$ programs the mean $\mu_i$ of weight $i$, whereas the optical bandwidth $\Delta\nu_i$ programs its standard deviation $\sigma_i$. Spectrally separated channels contain distinct frequency components and therefore produce mutually uncorrelated fluctuations [12]. A single broadband source can thus encode and sample multiple probabilistic weights in parallel. The bandwidth dependence originates from temporal averaging at the electronic readout, as sketched in Fig. 2a. Chaotic light with an optical bandwidth $\Delta\nu$ has a coherence time $\tau_c \sim 1/\Delta\nu$, whereas the finite electrical bandwidth $B_{\mathrm{el}}$ defines an effective measurement interval $\tau_m \sim 1/B_{\mathrm{el}}$. For a narrow optical channel, the detector resolves the slowly varying power fluctuations. Increasing the optical bandwidth shortens the coherence time, so that each measurement averages over an increasing number of effective temporal coherence cells, $M_{\mathrm{eff}} \sim \Delta\nu/B_{\mathrm{el}}$. This averaging suppresses the relative fluctuations approximately as $M_{\mathrm{eff}}^{-1/2}$. Optical power therefore sets the signal mean, whereas optical bandwidth controls the relative stochasticity.

We experimentally demonstrate this encoding using amplified spontaneous emission from an erbium-doped fibre source (Fig. 2b). A programmable waveshaper defines spectral channels with independently controlled power and bandwidth. We detect the resulting power fluctuations with a 38 GHz photodetector and record them using a real-time oscilloscope with a bandwidth of 33 GHz at 80 GSa $s^{-1}$. At fixed optical power, increasing the channel bandwidth progressively reduces the measured standard deviation because the detector averages over more temporal coherence cells (Fig. 2c). At large bandwidths, the standard deviation follows the expected $1/\sqrt{\Delta\nu}$ dependence before approaching the electronic-noise floor. At fixed bandwidth, increasing the optical power increases the absolute fluctuation amplitude approximately linearly. We extract $M_{\mathrm{eff}}$ from the measured voltage distributions at each operating point using a gamma distribution for the optical power fluctuations [16], a Gaussian distribution for the electronic ground noise and subtracting the mean signal due to the high-pass of the photodetector, also see the Supplementary Information[17]. The effective number of temporal coherence cells increases approximately linearly with optical bandwidth and remains largely independent of optical power as shown in Fig. 2d. The optical bandwidth also impacts the shape of the encoded distribution. When only a small number of coherence cells contributes to each measurement, the distribution retains the asymmetry of chaotic-light power fluctuations. Increasing the optical bandwidth averages over more independent fluctuations and produces a narrower, more Gaussian-like distribution. Representative measurements in Fig. 2e demonstrate both control dimensions. At a fixed bandwidth of 50 GHz, increasing the optical power from 25 µW to 75 µW increases the standard deviation from 0.010 V to 0.029 V. At a fixed power of 75 µW, increasing the bandwidth from 50 GHz to 500 GHz narrows the distribution to a standard deviation of 0.012 V and makes it more Gaussian-like.

These measurements establish optical power and bandwidth as programmable physical parameters for encoding probabilistic weights. The chaotic optical field stores the distribution parameters and continuously generates new stochastic realizations within the same physical signal.

## Photonic probabilistic convolution accelerator

We next combine spectral weight encoding with frequency–time interleaving [11] to compute convolutions with probabilistic weights directly in the optical domain, as illustrated in Fig. 3a. A waveshaper programs each weight distribution in a separate frequency channel of the amplified-spontaneous-emission spectrum. A broadband electro-optic modulator (EOM) encodes the same temporal input waveform onto all channels, and a chip-integrated chirped waveguide Bragg grating (CWBG) delays neighbouring channels by one input symbol. This frequency-dependent delay aligns each spectrally encoded weight with the corresponding input value at the photodetector. The detector sums the aligned optical powers, so that each measured output is a stochastic realization of the convolution between the input data and the programmed probabilistic kernel. The processor therefore generates the weight samples and evaluates the convolution within the same physical signal.

We experimentally implement a high-speed convolution accelerator using nine frequency channels centred around 194 THz and separated by 403 GHz. An 8-bit arbitrary-waveform generator (AWG) operating at 80 GSa $s^{-1}$ drives the EOM and encodes each input value as a three-sample symbol with a duration of 37.5 ps. After modulation, a fibre polarization controller sets the input to the transverse-electric mode of the CWBG. An 8-bit real-time oscilloscope (RTO), also operating at 80 GSa $s^{-1}$ and synchronized with the AWG, records the photodetector output. A PyTorch-compatible digital interface controls spectral programming, waveform generation and data acquisition. The integrated CWBG is fabricated on a low-loss silicon-nitride platform (Fig. 3b). Its measured frequency-dependent delay is $-93.12$ ps $THz^{-1}$ (Fig. 3e). Together with the 403 GHz channel spacing, this slope produces a delay of 37.5 ps between adjacent channels, exactly matching one encoded symbol. Once the pipeline is filled, the accelerator produces one probabilistic convolution output per symbol, corresponding to 26.7 billion outputs second. The measured delay from the AWG output to the RTO input remains approximately 70 ns across the operating band. Integrating the dispersive element on chip avoids the long fibre delay lines used in earlier deterministic frequency–time-interleaved processors [11] and provides the required channel alignment within a compact optical path and thus with low latency.

We program all nine weight distributions simultaneously using iterative feedback. In each iteration, the accelerator evaluates a set of test convolutions from which we estimate the realized means and standard deviations. We then update the power and bandwidth of each frequency channel according to the difference between the measured and target distributions; the full procedure is described in the Supplementary Information [17]. This calibration programs the parameters of each distribution rather than individual stochastic samples. Once programmed, the chaotic optical field continuously generates new weight realizations without further spectral reconfiguration. Convolutional weight sharing also reuses the same programmed distributions across many input positions, amortizing the programming overhead over the subsequent high-speed operations.

We characterize the accelerator using 25 randomly generated probabilistic kernels. For each kernel, we evaluate a random input vector of length 1000 and acquire 1500 samples from every output distribution. The measured output means and standard deviations closely follow their target values across the accessible operating range, as shown in Fig. 3c,d. The errors, defined as measured minus target, have mean ± s.d. values of 0.016 ± 0.173 for the output mean and 0.036 ± 0.092 for the output standard deviation. To incorporate the photonic processor into Bayesian-network design, we represent it as a PyTorch-compatible probabilistic convolution operator. A differentiable physical model predicts the mean and standard deviation of the convolution output from the input data and programmed weight distributions, while calibrated stochastic perturbations capture the remaining deviations observed experimentally. This statistical abstraction avoids modelling every optical and electronic imperfection separately while retaining the processor behaviour relevant to probabilistic inference. We use the abstract operator during hardware-aware training and replace it with the physical accelerator during experimental prediction.

## Uncertainty-aware photonic artificial intelligence

We co-design a Bayesian Neural Network to match the computational characteristics of the photonic Bayesian accelerator. The architecture contains a single probabilistic convolutional layer and combines DenseNet-style skip connections [18] with a modified form of the depthwise-separable convolutions introduced in MobileNetV1 [19], as shown in Fig. S2. An architecture search identifies the network structure and probabilistic-layer placement that maximizes the predictive performance, also see the Supplementary Information [17]. We implement the probabilistic convolution as a PyTorch-compatible operator [20, 21], allowing the deterministic digital layers to interact directly with either a software representation or the physical photonic accelerator. During training, a differentiable surrogate models the programmed weight distributions as Gaussian distributions and predicts the mean and standard deviation of the convolution output. We use stochastic variational inference (SVI) to infer these distributions from the training data [22]. Straight-through estimators include non-differentiable hardware constraints, such as converter quantization, output clipping and calibrated computing errors, in the forward pass while preserving gradients during backpropagation. PyTorch and Pyro [23] provide automatic differentiation and probabilistic inference, implementation details are given in the Supplementary Information [17].

For experimental prediction, we replace the software surrogate with the physical accelerator. We evaluate each input in ten stochastic forward passes. In every pass, the accelerator evaluates the probabilistic convolutional layer optically using a new physical realization of the encoded weights, while the surrounding deterministic layers are evaluated digitally. Each pass produces one predictive sample in the form of a class-probability vector. Averaging these vectors gives the posterior-predictive distribution. We calculate the Shannon Entropy H

$$\mathrm{H} = -\sum_{c=1}^{C}\left[\frac{1}{N}\sum_{n=1}^{N} p(y_n = c \mid x, \theta_n)\cdot \log\left(\frac{1}{N}\sum_{n=1}^{N} p(y_n = c \mid x, \theta_n)\right)\right] \tag{1}$$

from this averaged distribution and the softmax entropy SE

$$\mathrm{SE} = -\frac{1}{N}\sum_{n=1}^{N}\sum_{c=1}^{C} p(y_n = c \mid x, \theta_n)\cdot \log\big(p(y_n = c \mid x, \theta_n)\big) \tag{2}$$

as the mean entropy within the individual predictive samples. Their difference, MI=H-SE, is the mutual information. The softmax entropy captures uncertainty within individual predictions and therefore serves as a measure of aleatoric uncertainty, whereas the mutual information captures disagreement among predictions generated by different weight realizations and serves as a measure of epistemic uncertainty.

We first apply the hybrid photonic BNN to blood-cell classification. The training data contain seven classes: basophils, eosinophils, immature granulocytes, lymphocytes, monocytes, neutrophils and platelets, shown in Fig. 4a [24]. We withhold erythroblasts from training and introduce them as an out-of-distribution (OOD) class during prediction. SVI infers the probabilistic weights exclusively from the seven training classes, implementation details are provided in the Supplementary Information [17]. The posterior standard deviations of representative weights evolve during training and converge to distinct values, illustrating how the network learns the stochasticity of the probabilistic convolutional layer (Fig. 4b). The experimental predictions reveal different behaviour for in-distribution and withheld images. For a representative platelet image, the network produces consistent class probabilities across all ten stochastic forward passes and therefore negligible mutual information (Fig. 4e). For a withheld erythroblast image, the predicted class changes across weight realizations, producing high mutual information and indicating epistemic uncertainty (Fig. 4f). We therefore use mutual information to detect withheld erythroblast images. Varying the rejection threshold yields an area under the receiver operating characteristic curve (AUROC) of 91.16 % shown in Fig. 4c. At a mutual-information threshold of 0.0185, rejecting uncertain images increases the classification accuracy among retained in-distribution images from 90.26 % to 94.62 %. Figure 4d shows the corresponding confusion matrix for in-distribution and OOD images and the predicted labels or rejections.

Apart from detecting uncertainty in safety critical tasks, the hybrid photonic BNN further enables uncertainty reasoning, distinguishing epistemic novelty from aleatoric ambiguity. We train the network exclusively on MNIST and introduce two additional datasets only during inference, shown in Fig. 5a. Fashion-MNIST provides semantically unfamiliar OOD images associated with epistemic uncertainty, whereas Ambiguous-MNIST provides images that are intrinsically compatible with more than one digit class and therefore contain aleatoric uncertainty[15]. Unlike protocols that include ambiguous images during training [15, 25, 26], this procedure exposes the network to neither ambiguous nor OOD examples while it learns the weight distributions, faithfully mimicking real-world settings. Across ten predictive samples, a representative MNIST image produces consistent and confident class probabilities (Fig. 5b). A Fashion-MNIST image instead produces disagreement among predictive samples and consequently high mutual information (Fig. 5c). For an Ambiguous-MNIST image, the class-probability distribution remains broad within the individual predictive samples, producing high softmax entropy (Fig. 5d). The three datasets consequently occupy largely distinct regions of mutual-information–softmax-entropy space, separating epistemic novelty from aleatoric ambiguity (Fig. 5e). The experimentally hybrid photonic BNN achieves an MNIST classification accuracy of 96.01 %. Mutual-information-based rejection at a threshold of 0.003 08 increases the accuracy among non-rejected MNIST images to 99.70 %. Epistemic-uncertainty detection using Fashion-MNIST achieves an AUROC of 84.42 %, whereas aleatoric-uncertainty detection using Ambiguous-MNIST achieves an AUROC of 88.03 % (Fig. 5f).

Together, these experiments show that physically sampled probabilistic convolutions support both uncertainty-aware rejection and the separation of epistemic and aleatoric uncertainty in a hybrid photonic Bayesian neural network.

## Discussion

Spectral encoding turns broadband chaotic light into a programmable representation of probabilistic weights and thereby extends photonic uncertainty processing from stochastic elements to a complete probabilistic convolution accelerator. Previous studies proposed photonic Bayesian neural networks based on programmed optical noise, demonstrated stochastic photonic neurons or binary probabilistic states, and used experimentally characterized photonic devices to simulate Bayesian-network models [13, 27, 28, 14, 12]. Here, the optical power and bandwidth of each spectral channel program the mean and standard deviation of a weight distribution, while the same channel continuously generates new physical realizations and participates in the convolution. A PyTorch-compatible interface embeds this operation directly in an AI workflow and replaces the software operator with the physical accelerator during experimental prediction. To our knowledge, this combination provides the first experimental photonic implementation that unifies programmable probabilistic weights, optical convolution and end-to-end uncertainty-aware inference. The accelerator produces one probabilistic convolution output every 37.5 ps. Although rate definitions differ between architectures, this operating interval is substantially shorter than the nanosecond-scale stochastic operation reported or projected for other photonic probabilistic systems [27, 28]. Spectral encoding therefore advances the state of the art not only by accelerating random sampling, but by storing, sampling and processing probabilistic weights within the same optical signal.

The system-level value of the accelerator follows from the observation that Bayesian neural networks do not require every model parameter to be stochastic. Partially stochastic networks can retain expressive predictive distributions and match or outperform fully stochastic models while reducing their memory and computational requirements [29]. We exploit this principle through hardware–algorithm co-design. An architecture search concentrates stochasticity in one probabilistic convolutional layer and favours grouped and depthwise-separable operations that maximize predictive performance for a given number of photonic kernels. The nine simultaneously encoded weights form a complete $3 \times 3$ probabilistic kernel, while convolutional weight sharing reuses each programmed distribution across many spatial positions. The physical resource requirement therefore depends on the number of concurrently evaluated weights rather than on the total number of convolution operations. A differentiable abstraction of the accelerator further enables digital training without repeated access to the physical system. This approach follows the broader principle of hardware-aware training, in which calibrated device behaviour and non-idealities are represented in software and included during optimization [30]. During experimental prediction, we replace the abstract operator with the photonic accelerator and observe close agreement between their predictive performance. The resulting hybrid network detects withheld blood-cell classes and improves the accuracy of retained in-distribution images, while evaluation on the Dirty-MNIST uncertainty benchmark separates epistemic novelty from aleatoric ambiguity [24, 15]. The selective stochasticity, weight reuse and hardware-aware training make the photonic probabilistic convolution accelerator a practical system component for uncertainty-aware AI workloads.

The photonic probabilistic convolution accelerator offers a practical route towards full integration and larger-scale operation. The integrated chirped waveguide Bragg grating already replaces kilometre-scale fibre delays with a compact on-chip dispersive element, substantially reducing optical propagation time and latency compared with earlier frequency–time-interleaved processors [11]. The remaining optical functions can likewise be integrated. Erbium-doped silicon-nitride waveguides or electrically pumped gain media can provide broadband entropy sources [31, 32], while integrated electro-optic modulators and photodetectors operate at bandwidths exceeding 100 GHz [33, 34]. Dense wavelength-parallel processors and integrated spectral shapers further provide a route to programming more frequency channels with lower reconfiguration overhead [35, 36, 37]. Combining these components in an end-to-end integrated system would further reduce footprint and latency.

Our results demonstrate that spectrally encoded probabilistic weights transform intrinsic optical noise into a programmable computational resource, enabling high-speed uncertainty reasoning and establishing a path towards uncertainty-aware photonic artificial intelligence.

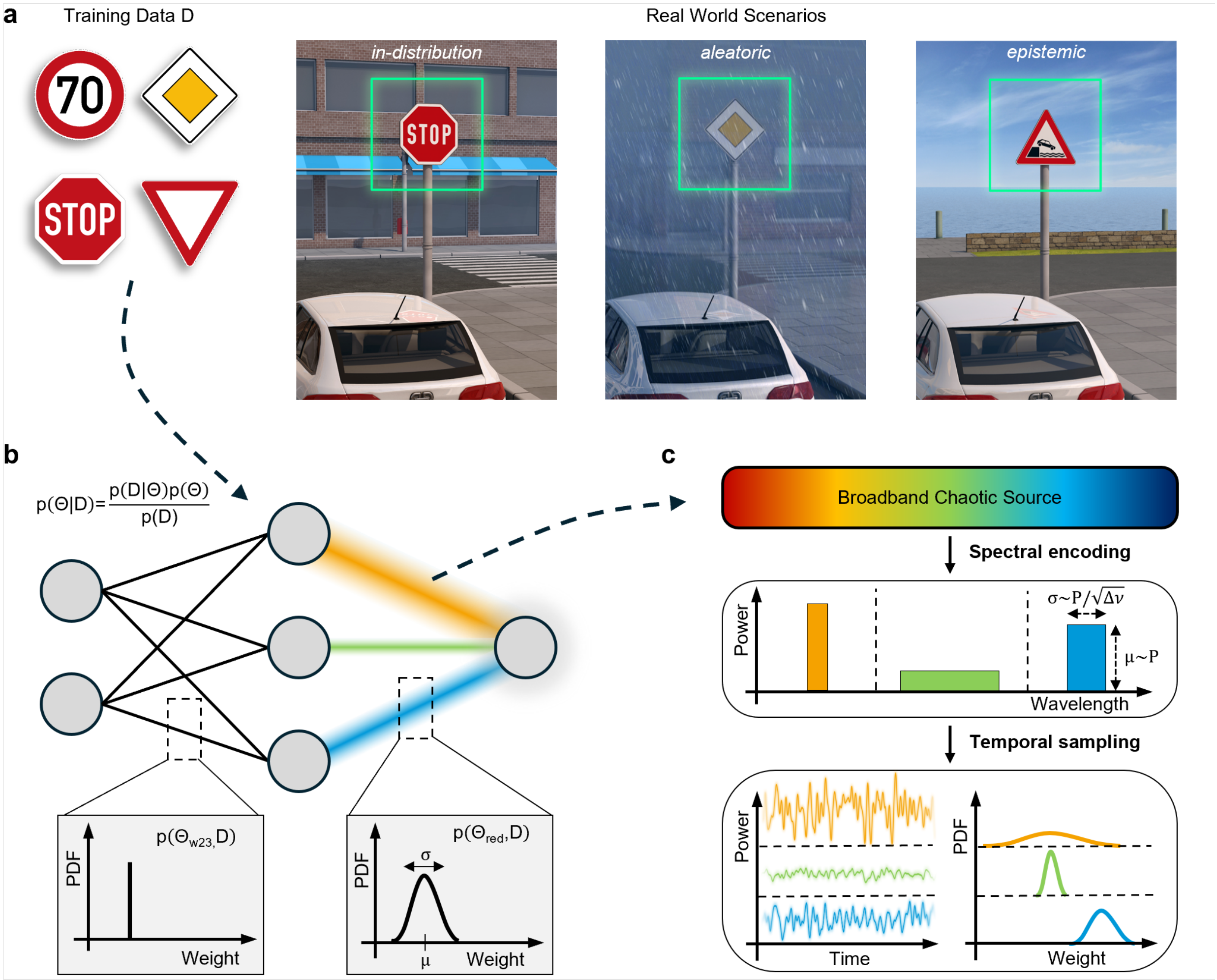

**Fig. 1: Uncertainty-aware photonic artificial intelligence.** **a**, In real-world applications, AI systems encounter aleatoric uncertainty, for example due to limited vision, and epistemic uncertainty arising from completely unknown inputs. Quantifying these uncertainties is crucial for deploying AI in safety-critical environments. **b**, Bayesian neural networks infer probability distributions over model parameters from the training data, whereas conventional networks infer single point estimates. This probabilistic formulation provides a principled framework for uncertainty reasoning. Only a subset of network weights needs to be stochastic [29]. Inference requires sampling from these distributions and repeating the forward pass multiple times, which incurs substantial computational overhead. **c**, We encode the weight distributions in the emission spectrum of a broadband chaotic light source and map each distribution to an individual frequency channel. Random beating between the frequency components within each channel produces temporal power fluctuations. We program the mean of the fluctuations by setting the total power $P$ in the frequency channel and program the standard deviation by setting the optical bandwidth $\Delta\nu$ of the channel. In this way, a single source both stores and samples from multiple weight distributions in parallel.

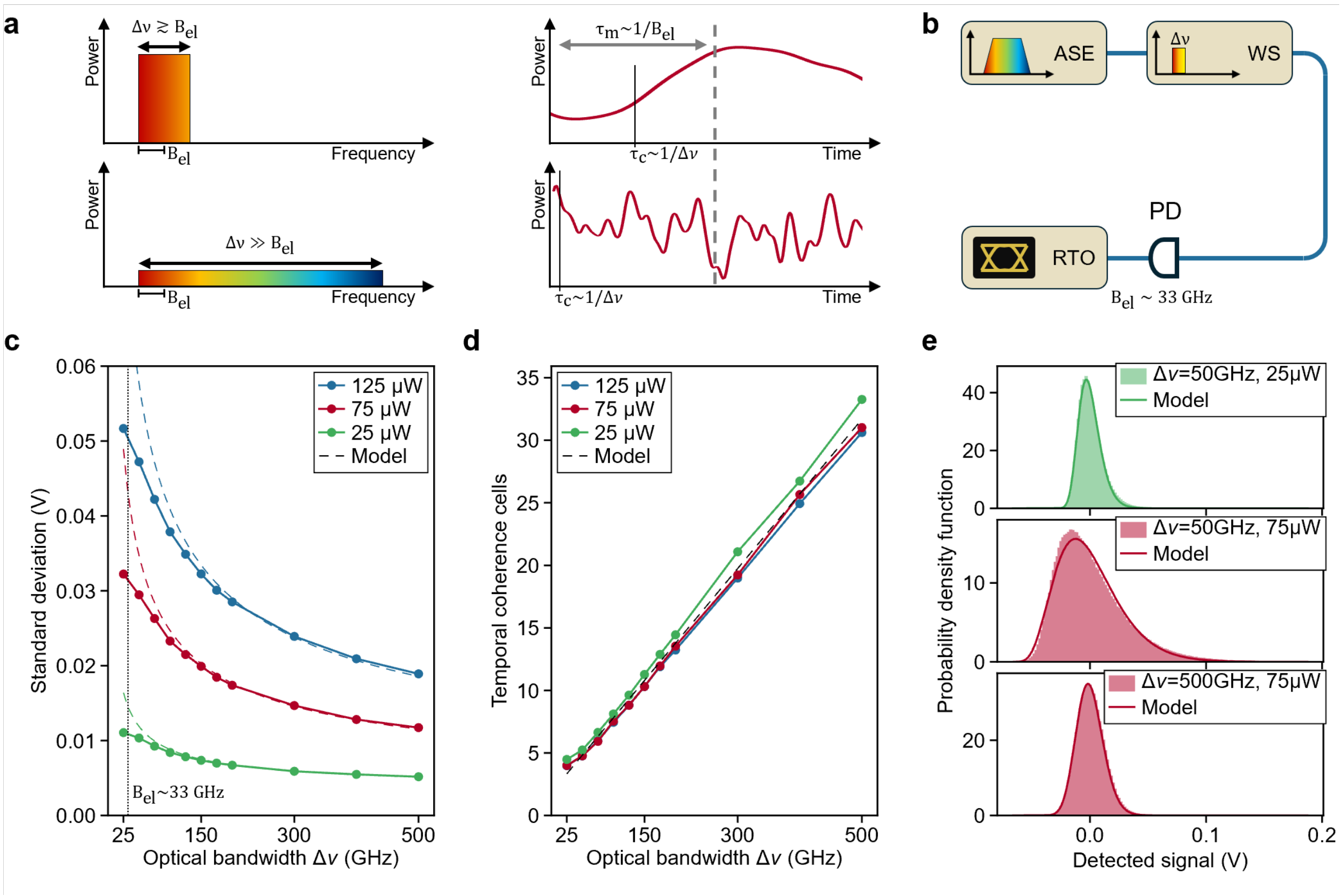

**Fig. 2: Spectral encoding of probability distributions.** **a**, Random beating between the frequency components of chaotic light produces temporal power fluctuations with a coherence time $\tau_c \sim 1/\Delta\nu$, where $\Delta\nu$ is the optical bandwidth. The finite electrical bandwidth $B_{el}$ of the electronic detection and readout system defines an effective measurement interval $\tau_m \sim 1/B_{el}$. For a narrow optical channel, the detector resolves the slowly varying fluctuations. Increasing the optical bandwidth shortens $\tau_c$, so that each measurement averages over multiple effective temporal coherence cells $M_{eff}$ and suppresses the detected fluctuations. **b**, We generate chaotic light through amplified spontaneous emission (ASE) in an erbium-doped fibre, program its spectrum with a waveshaper (WS), and record the power fluctuations using a 38 GHz photodetector (PD) and an 80 GSa s$^{-1}$ real-time oscilloscope (RTO) with a bandwidth of 33 GHz. **c**, We record RTO voltage traces at different optical powers and bandwidths. Increasing the optical bandwidth reduces the resolved fluctuations because the detector averages over more temporal coherence cells. At large optical bandwidths, the standard deviation follows $1/\sqrt{\Delta\nu}$ and eventually approaches the electronic-noise floor. Increasing the optical power increases the optical fluctuations linearly. **d**, We extract $M_{eff}$ from the measured voltage distributions at each operating point. $M_{eff}$ increases approximately linearly with optical bandwidth and remains largely independent of optical power, identifying bandwidth as the principal control parameter for temporal averaging. **e**, Representative measured probability densities. At a fixed bandwidth of 50 GHz, increasing the optical power from 25 µW to 75 µW proportionally increases the standard deviation from 0.010 V to 0.029 V. At a fixed power of 75 µW, increasing the bandwidth from 50 GHz to 500 GHz narrows the distribution to a standard deviation of 0.012 V and makes it more Gaussian-like.

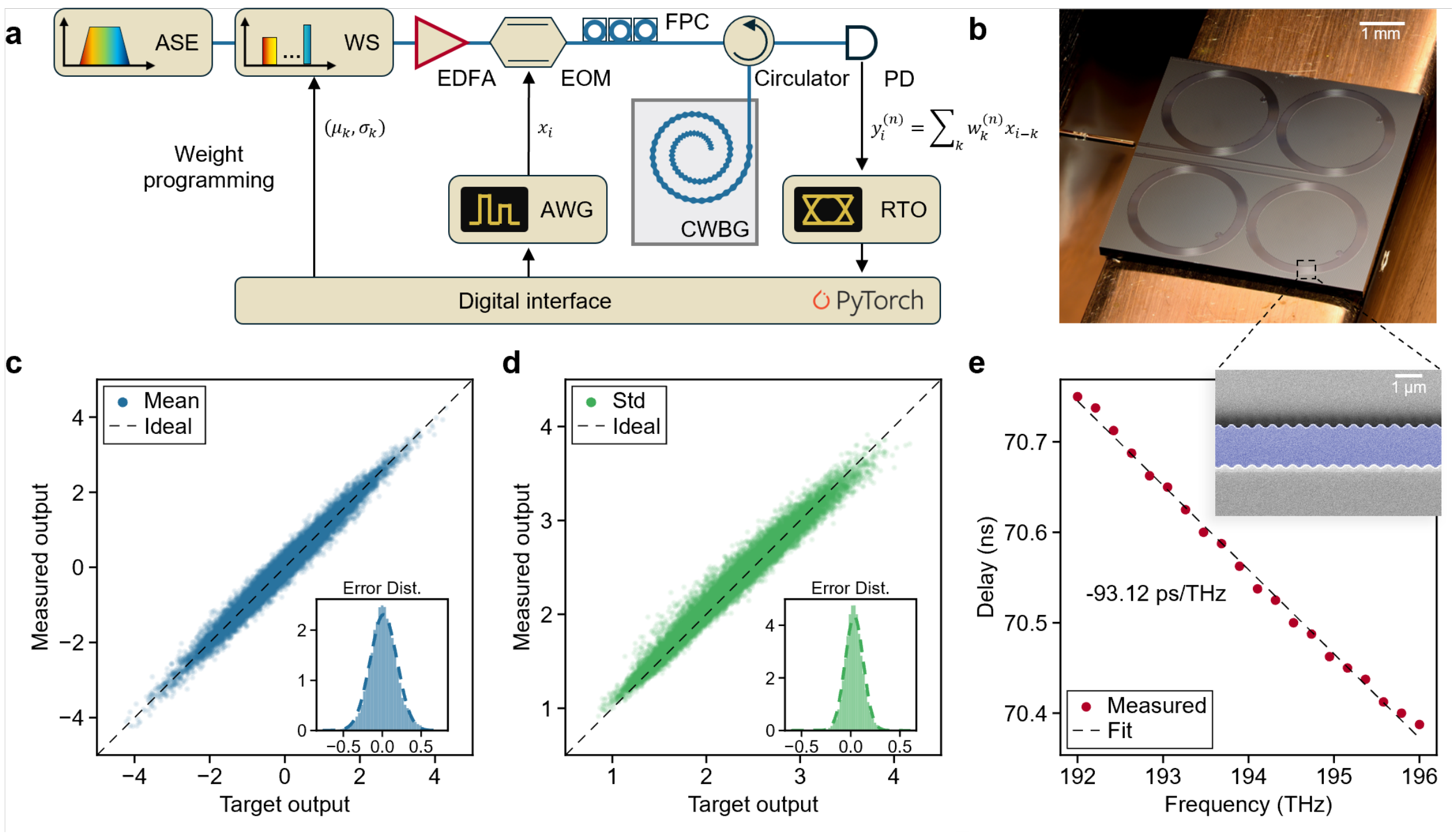

**Fig. 3: High-speed photonic convolution processing with probabilistic weights.** **a**, We combine spectral weight encoding with frequency–time interleaving to realize a high-speed photonic probabilistic convolution accelerator. A waveshaper (WS) programs nine probabilistic weights in frequency channels separated by 403 GHz within the emission spectrum of an amplified spontaneous emission (ASE) source. An 8-bit arbitrary-waveform generator (AWG) operating at 80 GSa $s^{-1}$ drives an electro-optic modulator (EOM), which encodes the temporal input data simultaneously onto all channels. Each input value is represented by a three-sample symbol with a duration of 37.5 ps. A chirped waveguide Bragg grating (CWBG) delays neighbouring channels by one symbol and aligns their contributions at the photodetector to implement a probabilistic convolution. A fibre polarization controller (FPC) sets the input polarization to the transverse-electric mode. An 8-bit real-time oscilloscope (RTO) operating at 80 GSa $s^{-1}$ records the output, while a PyTorch-compatible digital interface controls weight programming, waveform generation and data acquisition. **b**, Photograph of the integrated CWBG on a low-loss silicon-nitride platform. The on-chip dispersive element provides the frequency-dependent delay required for convolution while maintaining a short optical path. **c**,**d**, We characterize the accelerator using 25 randomly generated probabilistic kernels. For each kernel, we evaluate a random input vector of length 1000 and acquire 1500 samples from each output distribution. The measured output means (**c**) and standard deviations (**d**) closely follow their target values. The errors, defined as measured minus target, have mean ± s.d. values of 0.016 ± 0.173 for the output mean and 0.036 ± 0.092 for the output standard deviation. **e**, We measure the delay between the AWG output and the RTO input as a function of optical frequency. A linear fit yields a frequency-dependent delay of −93.12 ps $THz^{-1}$, corresponding to a delay of 37.5 ps between adjacent frequency channels. Inset, scanning electron micrograph of the chirped grating; its period varies along the propagation direction so that the centre of the reflection band shifts along the grating.

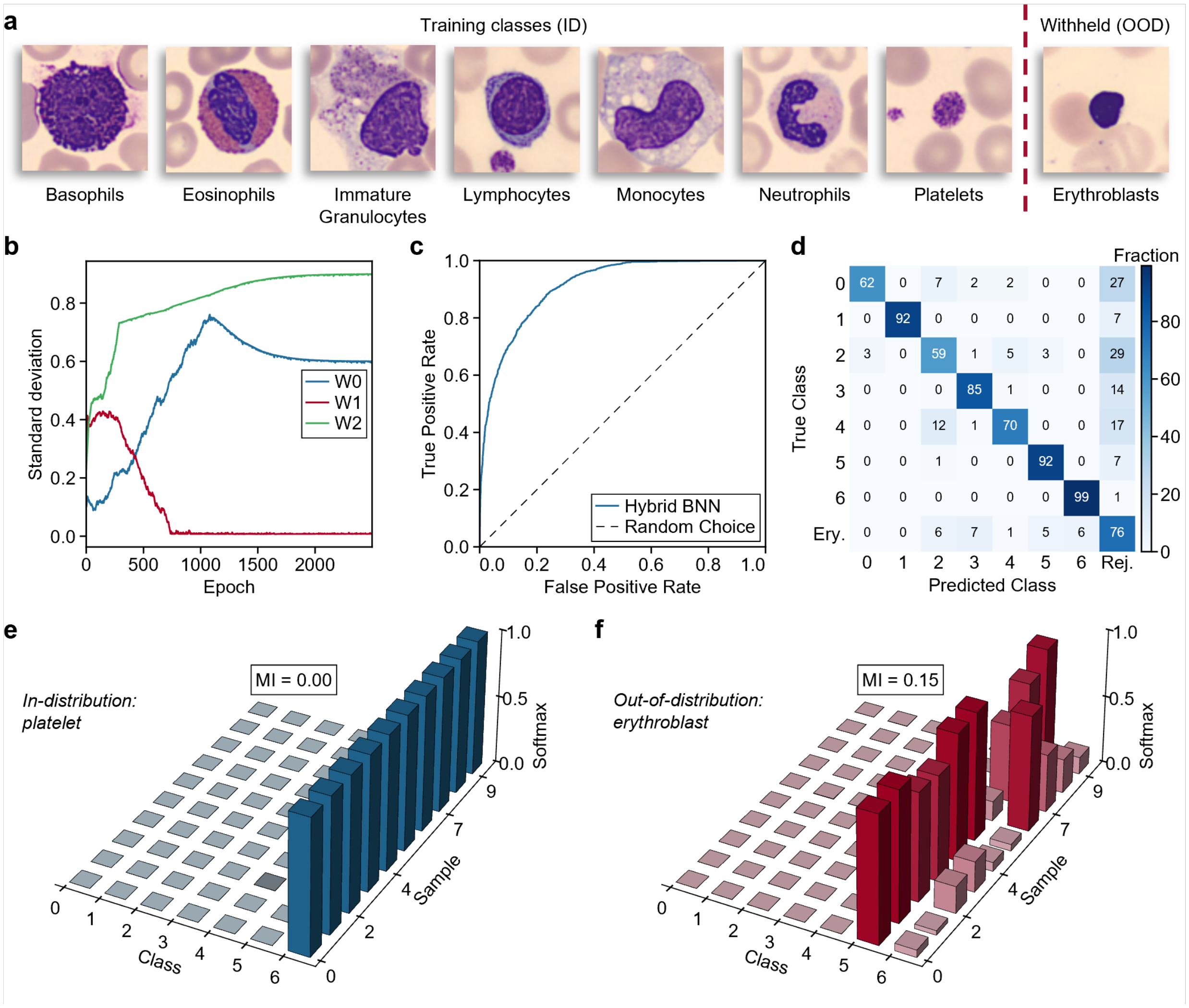

**Fig. 4: Uncertainty-aware blood-cell classification.** **a**, Representative microscope images from the seven blood-cell classes used for training. We withhold erythroblasts from training and introduce them as an out-of-distribution (OOD) class during inference. **b**, Evolution of the posterior standard deviations of three representative weights in the probabilistic convolutional layer during stochastic variational inference. **c**, Receiver operating characteristic (ROC) curve for detecting withheld erythroblast images using predictive mutual information (MI), yielding an area under the ROC curve (AUROC) of 91.16 %. The dashed line denotes random choice. **d**, Row-normalized confusion matrix at an MI rejection threshold of 0.0185. The final row denotes withheld erythroblast images, and the final column denotes rejected samples. At this MI threshold, the classification accuracy among retained in-distribution images increases from 90.26 % to 94.62 %. **e**,**f**, Class probabilities from ten stochastic forward passes of the hybrid photonic BNN for a representative in-distribution platelet image (**e**) and a withheld erythroblast image (**f**). In each pass, we evaluate the probabilistic convolutional layer optically using a new physical realization of the encoded weights and evaluate the surrounding deterministic layers digitally. The in-distribution image yields consistent predictions and negligible MI, whereas disagreement among predictions for the erythroblast produces high MI.

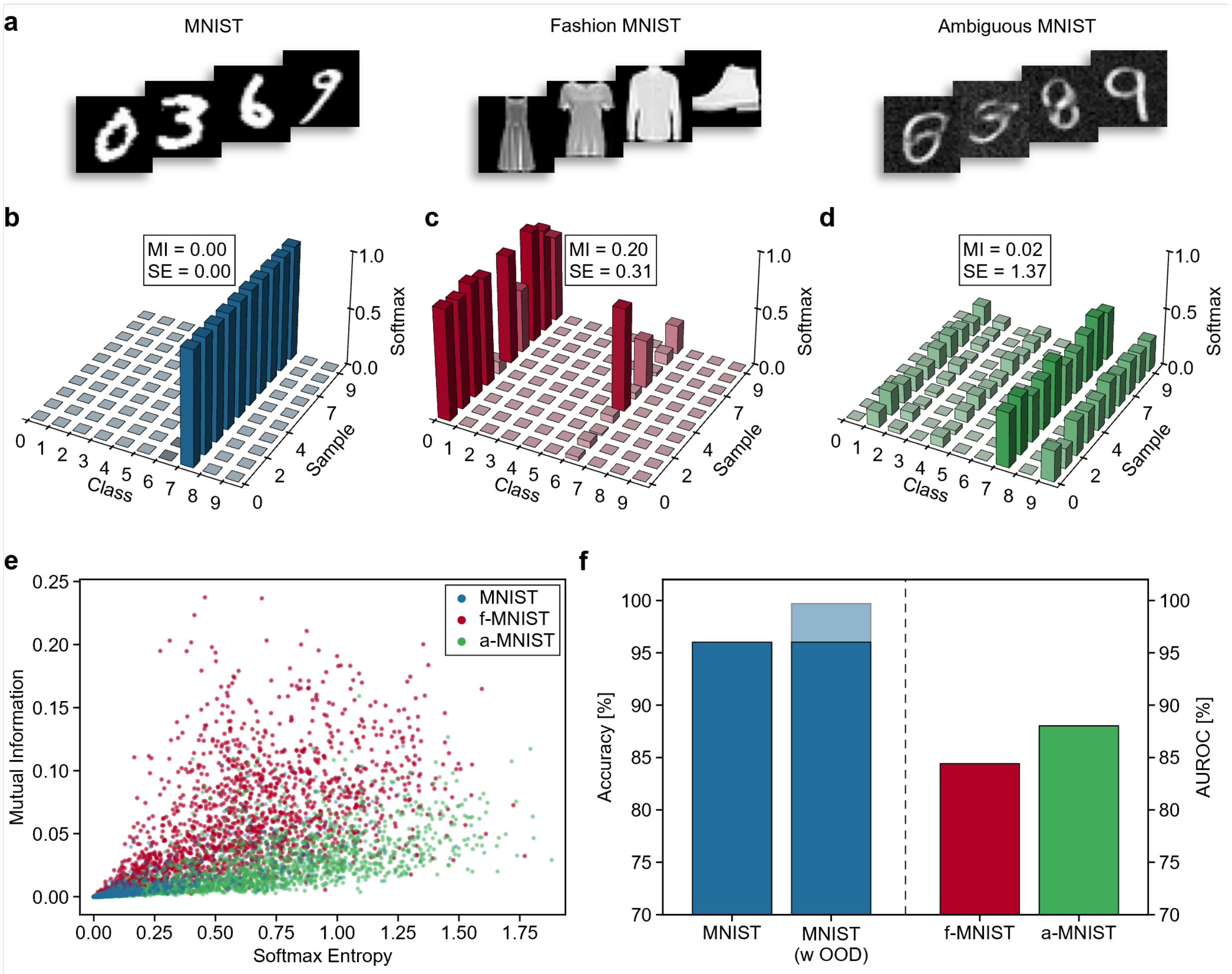

**Fig. 5: Disentanglement of aleatoric and epistemic uncertainty.** **a**, Representative images from MNIST, Fashion-MNIST and Ambiguous-MNIST. We train the hybrid photonic BNN exclusively on MNIST. Fashion-MNIST provides out-of-distribution images associated with epistemic uncertainty, whereas Ambiguous-MNIST provides intrinsically ambiguous images associated with aleatoric uncertainty. **b–d**, Class probabilities from ten predictive samples of the hybrid photonic BNN for representative MNIST (**b**), Fashion-MNIST (**c**) and Ambiguous-MNIST (**d**) images. Each sample corresponds to one stochastic forward pass with a new physical realization of the probabilistic weights. We perform all probabilistic convolutions optically and compute all other deterministic operations digitally. The in-distribution image yields consistent, confident predictions across samples. The out-of-distribution image produces disagreement among samples and therefore high mutual information (MI); the ambiguous image produces high entropy within individual samples, reflected in high softmax entropy (SE). **e**, Mutual information versus softmax entropy for individual test images. The three datasets occupy distinct regions of uncertainty space, enabling the separation of epistemic novelty from aleatoric ambiguity. **f**, Experimental task performance of the hybrid photonic BNN. Mutual-information-based rejection at a threshold of 0.003 08 increases the MNIST classification accuracy from 96.01 % to 99.70 % for non-rejected test images. Epistemic- and aleatoric-uncertainty detection achieve AUROCs of 84.42 % for Fashion-MNIST and 88.03 % for Ambiguous-MNIST, respectively.

## Materials and Methods

### Experimental Setup

We house the photonic accelerator and its optical and electronic subsystems in a standard 19-inch rack. An Agiltron ASES-1611A3113 amplified-spontaneous-emission source generates broadband chaotic light. A Coherent WaveShaper 1000B programs the probabilistic weights by setting the power and bandwidth of the individual frequency channels. We subsequently amplify the shaped spectrum using the preamplifier stage of a PriTel LNTFA-20-NMA-DP erbium-doped fibre amplifier, which increases the optical power with low excess noise. An Optilab IML-1550-40-PM-V electro-optic modulator encodes the analogue electronic input waveform onto the shaped optical spectrum.

The modulated signal passes through a fibre polarization controller and an optical circulator before entering the integrated chirped waveguide Bragg grating (CWBG) shown in Fig. 3). The polarization controller sets the input to the transverse-electric mode of the CWBG, which introduces a frequency-dependent delay of 37.5 ps between adjacent frequency channels. The circulator separates the reflected convolution signal from the optical input. A Thorlabs RXM38AF photoreceiver detects the summed optical output. The photoreceiver comprises a high-speed photodiode with a responsivity of 0.7 $\mathrm{A\,W^{-1}}$ and a tunable-gain transimpedance amplifier operated at a total conversion gain of 1000 $\mathrm{V\,W^{-1}}$.

A Keysight M8196A arbitrary-waveform generator (AWG) provides the digital-to-analogue interface, and a Keysight DSAX95004Q real-time oscilloscope (RTO) provides the analogue-to-digital interface. Both instruments operate at 80 $\mathrm{GSa\,s^{-1}}$. We use pulse-amplitude modulation with three samples per symbol, corresponding to a symbol duration of 37.5 ps. A Xilinx ZCU216 board generates phase-locked 100 MHz and 2.5 GHz reference clocks that synchronize the AWG and RTO and align their sampling rates. Python scripts control the waveshaper, AWG and RTO and expose the photonic accelerator through a PyTorch-compatible interface.

### Chirped Grating Fabrication

We fabricate the CWBGs on a wafer-scale, 200 nm-thick $Si_3N_4$ platform [38]. Fabrication starts from a 4-inch silicon wafer with a 10 µm wet-oxide $SiO_2$ layer that forms the lower cladding. We deposit a 200 nm $Si_3N_4$ layer by low-pressure chemical-vapour deposition and anneal it at 1200 °C to reduce hydrogen-related optical loss. We define the waveguides by deep-ultraviolet lithography (KrF 248 nm) and transfer the pattern by anisotropic reactive-ion etching with fluorine-based gases. We then encapsulate the waveguides in a 3 µm hydrogen-free, low-loss $SiO_2$ upper cladding [39]. We separate the dies by deep reactive-ion etching through the oxide and partially into the silicon substrate, followed by backside grinding.

The CWBG contains 113 000 grating periods that vary linearly from 496.7 nm to 509.3 nm. The resulting 56.8 mm-long device is folded into a spiral to reduce its footprint. We choose a sufficiently large spiral radius to limit radius-dependent variations in the effective refractive index [40]. The waveguide has an average width of 1.6 µm, close to single-mode operation, and a periodic width corrugation that provides distributed Bragg reflection. A photograph of the integrated CWBG is shown in Fig. 3b, and the inset of Fig. 3e shows a scanning electron micrograph of the corrugated waveguide.

We suppress reflections from the grating boundaries by apodizing the corrugation amplitude with a hyperbolic-tangent profile [40, 41]. Sharp waveguide bends terminate the distal end of the CWBG and further suppress end-facet reflections. We simulate the optical response using the transfer-matrix method [42], representing each grating period by a $2 \times 2$ transfer matrix. To reduce the computational cost, we divide the grating into groups of 100 consecutive periods and assume identical matrices within each group. Figure 3e shows the measured frequency-dependent delay of the fabricated device. We package the CWBG with an ultra-high-numerical-aperture silica fibre (UHNA7) for integration into the rack-mounted photonic setup.

## Neural Network Training

We implement and train the Bayesian neural networks in PyTorch [20, 21]. Pyro provides stochastic variational inference (SVI) [23], and Brevitas provides quantization-aware training [43]. We combine these frameworks in a custom, SVI-compatible probabilistic convolution operator that represents the photonic accelerator during training. The operator models the physical convolution statistics together with relevant hardware constraints, including digital-to-analogue converter quantization, the analogue-to-digital converter range and stochastic deviations in weight programming. We use the tensor-manipulation library Einops to implement the custom depthwise-separable (DWS) convolutions and associated tensor rearrangements [44].

We train the networks on a heterogeneous compute cluster managed with SLURM [45]. The available CPUs comprise AMD EPYC 7452, AMD EPYC 7302P and Intel Xeon E5-2667 v3 processors. The corresponding GPU resources comprise NVIDIA A30, NVIDIA GeForce RTX 2080 Ti and NVIDIA Tesla K80 devices. We manage and document the training runs using the open-source SEML framework, which stores configurations and results in an on-premises MongoDB database. A custom SEML extension implements the asynchronous successive-halving algorithm (ASHA) as an early-stopping procedure for efficient hyperparameter optimization [46].

During experimental inference, we map the two-dimensional probabilistic convolutions used by the network onto the hardware-compatible one-dimensional convolutions evaluated by the photonic accelerator. Custom data-reordering functions transform the input tensors and programmed kernels before optical execution and restore the two-dimensional output layout afterwards. We compile these functions just in time using Numba [47]. This implementation maximizes data reuse while maintaining a common PyTorch interface for software training and physical prediction.

### Acknowledgments

We thank Jochen Stuhrmann, from Illustrato, for his assistance with the illustrations. The samples were partially fabricated in the EPFL Center of MicroNanoTechnology (CMi).

**Funding:** This work was supported by funding from the European Research Council grant 101200429 (WP) and Contract N660012424006 (NaPSAC) from the Defense Advanced Research Projects Agency (DARPA), Defense Sciences Office (DSO) (TK, WP).

**Author contributions:** Conceptualization: FBP, HB, LM, BK
Methodology: FBP, JH, SH, XJ, FE, JD, LM, PS, JS
Investigation: FBP, HB, LM, JH
Visualization: FBP, HB, SH, XJ
Funding acquisition: TK, HF, WP
Project administration: FBP, TK, HF, WP
Supervision: TK, HF, WP
Writing – original draft: FBP, HB
Writing – review & editing: All authors

**Competing interests:** There are no competing interests to declare.

**Data and materials availability:** All data are available in the main text or the supplementary materials.

### Supplementary materials

Supplementary Text
Figs. S1 to S6
Table S1

# Supplementary Materials for

## Temporal speckle enables probabilistic weights for uncertainty-aware photonic AI

F. Brückerhoff-Plückelmann[1*], H. Borras[2*], S. U. Hulyal[3*], L. Meyer[1], X. Ji[3], J. Hu[3], J. Sun[3], B. Klein[2], F. Ebert[1], J. Dijkstra[1], L. McRae[1], P. Schmidt[1], T. J. Kippenberg[3†], H. Fröning[2], W. Pernice [1†]

[*]These authors contributed equally to this work.

[†]Corresponding author. Email: wolfram.pernice@kip.uni-heidelberg.de, tobias.kippenberg@epfl.ch

## Contents

## Supplementary Text

### Probability density of the measured optical fluctuations

Amplified spontaneous emission produces a chaotic optical field whose frequency components have random relative phases. Square-law photodetection converts their mutual beating into temporal power fluctuations. We consider a temporally varying input sequence encoded by the electro-optic modulator. The input sequence has zero temporal mean, such that, within the linear modulation regime, the time-averaged detector signal remains constant over the acquisition window.

For a fixed input value, repeated realizations of the detector signal before high-pass filtering follow a gamma distribution,

$$f(z;\mu,M) = \frac{M^M}{\mu\,\Gamma(M)}\left(\frac{z}{\mu}\right)^{M-1}\exp\!\left(-\frac{Mz}{\mu}\right), \qquad z \geq 0, \tag{S1}$$

where $\mu$ is the conditional mean signal before high-pass filtering and $M$ is the effective number of independent temporal coherence cells contributing to one electrical measurement [16]. The parameter $M$ is not restricted to integer values. This parameterization gives $\mathrm{Var}(z) = \mu^2/M$, such that the relative standard deviation decreases as $M^{-1/2}$.

The photoreceiver is AC-coupled and removes the slowly varying component $\Delta x$ of the detector signal. We denote the measured signal after high-pass filtering by $x$ and its conditional mean by $\bar{x}$. The corresponding pre-filter mean is therefore $\mu = \bar{x} + \Delta x$, and the probability density of the optical fluctuations after high-pass filtering is

$$f_{\mathrm{PD}}(x;\bar{x},M,\Delta x) = f(x+\Delta x;\bar{x}+\Delta x, M)\,, \qquad x \geq -\Delta x. \tag{S2}$$

For the zero-mean input encoding used here, $\Delta x$ corresponds to the constant detector contribution at the electro-optic-modulator bias point. The input-dependent component changes $\bar{x}$ but does not change the baseline removed by the high-pass filter.

The electronic readout additionally contributes independent, zero-mean Gaussian noise,

$$g(v;\sigma_{\mathrm{el}}) = \frac{1}{\sigma_{\mathrm{el}}\sqrt{2\pi}}\exp\!\left(-\frac{v^2}{2\sigma_{\mathrm{el}}^2}\right), \tag{S3}$$

where $\sigma_{\mathrm{el}}$ is the electronic-noise standard deviation. The measured photodetector voltage distribution is the convolution of the shifted gamma distribution and the electronic-noise distribution,

$$p(x;\bar{x},M,\Delta x,\sigma_{\mathrm{el}}) = \int_0^\infty f(z;\bar{x}+\Delta x, M)\, g(x-z+\Delta x;\sigma_{\mathrm{el}})\;\mathrm{d}z. \tag{S4}$$

For the measurements in Fig. 2, we apply no temporal input modulation. The optical statistics are therefore stationary and, once the high-pass transient has decayed, the measured distribution is centred around zero, such that $\bar{x} = 0$. The removed component $\Delta x$ then equals the mean of the stationary pre-filter signal. We determine $\sigma_{\mathrm{el}}$ from measurements with the optical input blocked and keep it fixed for all fits. Because we hold the integrated optical power constant while varying the channel bandwidth, we use a common $\Delta x$ for all bandwidths measured at the same optical power. We further constrain $\Delta x$ to scale linearly with the measured optical power across the three power settings. We fit $M$ independently at each operating point. The extracted values increase approximately linearly with optical bandwidth and remain largely independent of optical power, as shown in Fig. 2d.

### Physical Computation Model

The photonic accelerator implements a probabilistic one-dimensional convolution, but its analogue operation depends on several physical parameters. These include the bias-point transmission $b$ of the electro-optic modulator (EOM), which couples the analogue electronic input to the optical signal; the encoding range $r$ around the bias point, which depends on the $\pi$-voltage of the EOM and the applied electronic voltage swing; and the electronic-noise standard deviation $\sigma_{\mathrm{el}}$. During computation, the oscilloscope clips the detector voltage to $\pm 110\,\mathrm{mV}$ and maps this interval to the normalized output range $[-16, 16]$. We choose this range because the maximum mean output lies within $[-9, 9]$ for a convolution with inputs in $[-1, 1]$ and a nine-component kernel with weights in $[0, 1]$. We set the EOM bias-point transmission to $b = 0.4$ via the EOM calibration measurement.

We determine the encoding range $r$ and the electronic ground noise $\sigma_{el}$ by fitting the following computation model to measured output distributions with different optical powers and bandwidths for each frequency bin. For an input vector $x$ with $x_i \in [-1, 1]$ the mean optical power after the EOM at time step $i$ is:

$$\langle P_{\mathrm{EOM}} \rangle_i = \sum_{k=1}^{9} \langle P_{k,\mathrm{weight}} \rangle \left( b + r \cdot x_i \right) \tag{S5}$$

Here $\langle P_{\mathrm{k,weight}} \rangle$ is the mean optical power of each frequency bin. Since the insertion loss of all components is constant over time, it's already implicitly considered here.

After the chirped waveguide Bragg grating applies the frequency-dependent group delay, the mean optical power at the photodetector at time step $i + \tau$ is

$$\langle P_{\mathrm{PD}} \rangle_{i+\tau} = \mathrm{conv1d}\big(b + rx,\ \big\langle P_{\mathrm{weight}} \big\rangle\big)_i\,, \tag{S6}$$

where $\tau$ is the total sample delay between input sample $i$ in the arbitrary-waveform-generator waveform and the corresponding output sample in the oscilloscope waveform. Because the highest-frequency channel experiences the smallest group delay, we reverse the weight order when mapping the kernel coefficients to the frequency channels. We use the PyTorch convention for conv1d, which differs from the NumPy convolution convention by a reversal of the weight vector.

The analogue readout electronics applies a high-pass filter with a cut-off frequency of 45 kHz while amplifying the detector output. For the zero-mean input encoding used here, the high-pass filter removes the constant bias contribution from the mean measured voltage. We use a constant scaling factor $\alpha$ to account for the $1000\,\mathrm{V\,W^{-1}}$ conversion gain of the photodetector and transimpedance amplifier, together with the rescaling applied after analogue-to-digital conversion. The mean output distribution is therefore

$$\langle y \rangle = \mathrm{conv1d}\big(x,\ \alpha r \left\langle P_{\mathrm{weight}} \right\rangle\big)\,. \tag{S7}$$

The accelerator can thus implement convolution operations with signed inputs and positive weights by programming the optical power in each frequency channel.

Because the power fluctuations in different frequency channels are uncorrelated, their variances add at the photodetector. The variance of the output distribution is

$$\mathrm{Var}(y) = \mathrm{conv1d}\left(\left(x + \frac{b}{r}\right)^2,\ \left(\alpha r \sigma_{\mathrm{weight}}\right)^2\right) + \sigma_{\mathrm{el}}^2, \tag{S8}$$

where the squares are evaluated element-wise. Although the high-pass filter removes the bias-point contribution from the mean output, the bias still scales the high-frequency optical fluctuations and therefore contributes to the output variance. The standard deviation $\sigma_{\mathrm{weight}}$ of each probabilistic weight depends on the optical power and bandwidth of the corresponding frequency channel. We use $r = 0.26$ and $\sigma_{\mathrm{el}} = 0.52$ to emulate the physical accelerator during hardware-aware training.

### Error model

Hardware-aware training requires a robust and computationally efficient measure of analogue computation accuracy. We model the residual deviations of the physical accelerator by adding Gaussian-distributed perturbations to the output mean and standard deviation predicted by the floating-point computation model:

$$\begin{aligned} \bar{y}_{\mathrm{meas}} &= \bar{y}_{\mathrm{fp}} + \lambda_{\mathrm{mean}}\,\mathcal{N}(0,1), \\ \mathrm{std}(y_{\mathrm{meas}}) &= \mathrm{std}\big(y_{\mathrm{fp}}\big) + \lambda_{\mathrm{std}}\,\mathcal{N}(0,1). \end{aligned} \tag{S9}$$

The Gaussian perturbations are sampled independently for the output mean and standard deviation. To reproduce the experimentally observed hardware deviations during training, we use $\lambda_{\mathrm{mean}} = 0.11$ and $\lambda_{\mathrm{std}} = 0.053$.

### Delay and skew calibration

Accurate calibration of the temporal offset between the symbols transmitted by the digital-to-analogue converter (DAC) and those recorded by the analogue-to-digital converter (ADC) is essential for correct convolution alignment. Because the propagation delay depends on optical frequency, we perform the calibration at the system centre frequency of 194 THz. We disable all other frequency channels, such that the system implements an identity operation rather than a temporal convolution.

We first transmit a random input vector containing 80 000 components and determine the optimal integer sample delay at the ADC, corresponding to the number of samples between the oscilloscope trigger and the recorded output waveform. Figure S1a shows the deviation between the transmitted and received symbols as a function of sample delay and exhibits a clear minimum. The minimum extends across more than one sample because each symbol is represented by three samples, which relaxes the timing requirements.

We then calibrate the sub-sample delay by sweeping the oscilloscope skew, which controls the temporal offset between the trigger and ADC acquisition. We vary the skew in steps of 1 ps and determine the optimal sample delay at each setting (Fig. S1b). The resulting plateau-like behaviour reflects the 12.5 ps sample duration at the 80 GSa $s^{-1}$ acquisition rate. We set the skew to the centre of the plateau to maximize tolerance to residual timing errors.

#### Iterative weight programming

Rather than programming each probabilistic weight independently, we optimize all frequency channels simultaneously using an iterative feedback procedure. In each iteration, the photonic accelerator evaluates probabilistic convolutions for a random input vector of length 300, and we acquire 2500 realizations of each output distribution. We fit the physical computation model to the measured distributions to estimate the currently programmed weight means $\mu_\mathrm{p}$ and standard deviations $\sigma_\mathrm{p}$.

We then update the channel power $P$ and optical bandwidth $\Delta\nu$ according to their deviations from the target mean $\mu_\mathrm{t}$ and standard deviation $\sigma_\mathrm{t}$:

$$\begin{aligned} \Delta P &= \frac{\mu_\mathrm{t} - \mu_\mathrm{p}}{\alpha}, \\ \Delta(\Delta\nu) &= -\frac{\sigma_\mathrm{t} - \sigma_\mathrm{p}}{\beta\mu_\mathrm{p}}. \end{aligned} \tag{S10}$$

We apply these updates independently to each frequency channel. The calibration factors $\alpha$ and $\beta$ approximate the responses of the programmed mean to optical power and of the standard deviation to optical bandwidth, respectively.

We scale the updates using a cosine-annealed factor that decreases from 1.5 in the first iteration to 0.5 in the final iteration. We use ten iterations to program the complete probabilistic kernel. Throughout the procedure, we constrain the channel bandwidth to the range 25 GHz to 150 GHz to limit pulse-shape distortion arising from intrachannel dispersion.

### Stochastic variational inference

Stochastic variational inference (SVI) provides a natural framework for learning the parameterized probability distributions represented by the photonic accelerator. SVI approximates an analytically intractable posterior distribution with a tractable variational distribution whose parameters are optimized by gradient descent [48, 49]. This approach is compatible with the photonic hardware because the encoded weight distributions are parameterizable and their likelihoods can be evaluated efficiently.

We use the Bayes-by-Backprop formulation of SVI [6], which applies the reparameterization trick to express stochastic sampling as a deterministic transformation of the learned distribution parameters and an auxiliary noise variable. This formulation enables gradients to propagate through stochastic network parameters during backpropagation.

We construct the Bayesian neural network in the same manner as a deterministic deep neural network, using stacked fully connected and convolutional layers, also see Fig. S2. We replace selected deterministic parameters with probability distributions while retaining the remaining parameters as point estimates. Gradient-based optimization then jointly learns the deterministic parameters and the parameters of the probabilistic weights.

### Bayesian Neural Network Design

At the core of the Bayesian Neural Network implementation is the Probabilistic Programming Language: Pyro. The language is designed with SVI as its central component and provides a high-performance interface to GPU accelerated Bayesian inference, by building on top of PyTorch. Within Pyro we implement the custom probabilistic software component, modeling the photonic Bayesian machine, as a Pyro Module. Here we put the primary design focus on flexibility to adapt to the hardware as it changes throughout the co-design process. To support SVI for the photonic Bayesian machine the operator implements both Eq. S7 and Eq. S8 on a PyTorch level. While learning the deterministic weights required for Eq. S7 is trivial using the automatic gradient feature of PyTorch, we apply the following equation to arrive at $\sigma_{weight}$ to implement Eq. S8:

$$\sigma_{\text{weight}} = \left(\sigma_{\text{rel}} \cdot (\sigma_{\text{max}} - \sigma_{\text{min}}) + \sigma_{\text{min}}\right) \cdot \langle P_{\text{weight}} \rangle \tag{S11}$$

Here $\sigma_{rel}$ is the learnable probabilistic parameter, while $\sigma_{min}$ and $\sigma_{max}$ represent the bounds within which the hardware can represent the parameter. The advantage of this approach is that we do not need to account for changes to the hardware in the SVI learning process, as this is handled by flexibly setting the constant bound parameters. As $\sigma_{rel}$ is a learnable probabilistic parameter, we implement it as a Pyro parameter. A uniform prior bound between [0,1] effectively ensures that the posterior is bounded to the same range. We then utilize a posterior of randomly initialized Delta distributions, which change their parameters to approximate the true posterior during training. During training the expected Gaussian distribution of the hardware is sampled, which is computed from the results of Eq. S7 and Eq. S8. To model differences between the floating-point representation of the software and the actual hardware we primarily use straight-through gradient estimators. Their efficacy is well explored in the context of Quantization Aware Training [50]. From a basic functionality perspective, the idea is to apply disturbances (such as quantization) to the floating-point variables during the forward path of a neural network, while disregarding them during the backward path. This allows for the computation of the full gradient during the backwards path, even if the observed disturbances are generally non-differentiable. Using this principle we model the following hardware properties:

- Discretization of the DAC: quantization of fixed scale to eight bits (using QAT based on Brevitas)
- Clipping of the ADC: clamping of the operator output to: [-16, 16] (using a custom straight-through gradient estimator)
- Programming precision of the hardware for the parameters of $\sigma_{weight}$ and $\langle P_{\text{weight}} \rangle$, as described in Eq. S9, using a straight-through gradient estimator

To interface with the photonic hardware, we develop a PyTorch compatible interface. As the hardware executes a 1D convolution, the API is made to be compatible with the functional implementation of a 1D convolution in PyTorch. This simplifies development, as no access to the hardware is required during initial development, only for the later evaluation. To this end the custom software operator implements a second execution path to switch from software to hardware execution, without the need for changes to the BNN model. The hardware execution path in particular encompasses the conversion of the 2D convolution to a 1D convolution. To enable data reuse and fast execution times, we implement a custom conversion function with numba, which converts the inputs $\sigma_{weight}$ and $\langle P_{\text{weight}} \rangle$ from 2D to 1D and back. This is required, as such conversions are not possible with high data reuse within PyTorch. Here, we ensure high data reuse and efficient execution by making use of the just-in-time compilation capabilities of the software package numba. Similarly, after the hardware execution, the 1D outputs are converted back to a 2D tensor, for further processing within the rest of the model. The architecture design of the BNN requires careful consideration of the photonic Bayesian machine. As the hardware highly values grouped convolutions over standard ones, we deploy an architecture search to find BNN models most suitable for the photonic hardware. We in particular consider DenseNet-like skip-connections [18] to increase the robustness of the resulting architecture with respect to noise [51]. Along with this, we employ a custom type of Depthwise Separable convolutions as described by MobileNet [19]. This again maximizes the number of grouped convolutions. However, in contrast to standard DWS convolutions, we perform the pointwise convolution as a 1D convolution along the channel dimension. A survey of different architecture candidates is shown in Fig. S3, while the final resulting architecture is described in the main manuscript. As a result of hyperparameter optimization we also find an optimal activation function for the network, out of a selection of many. Notably this activation function is a modification to LeakyReLU, as proposed by Tempczyk et. al [52], where the slope of the negative domain is set to -0.5, which allows for more Gaussian-like posteriors and lower expected calibration errors.

### Bayesian Neural Network Training

After designing the BNN architecture as above, we train it using the standard SVI training loop of Pyro. Along with dataset preprocessing in the form of normalization for all datasets we also employ data augmentation for the MedMNIST dataset [24]. The official reference implementation of TrivialAugment [53] is used for data augmentation. This efficiently allows us to control the amount of data augmentation in a range of 0 to 32. Notably for the Dirty-MNIST [54] testcase we only train on MNIST, which is a conceptually more difficult task than the standard task [15, 25, 26], in which one trains on the training splits of both MNIST and Ambigious-MNIST at the same time. For our network the capability to distinguish aleatoric and epistemic uncertainty thus purely arises from the Bayesian Neural Network. For the MedMNIST task the Bayesian Neural Network is then trained with the following hyperparameters, which are the result of an extensive Hyperparameter search using ASHA:

- Training Dataset: bloodmnist
- Filtered_class: 2
- Data augmentation factor: 17
- train_batch_size: 256
- num_epochs: 2500
- num_train_samples: 3
- Loss function: TraceMeanField_ELBO
- learning rate: 0.000528
- weight_decay: 0.
- Optimizer: AdamW
- KL annealing schedule: linear
- KL max: 0.327
- Activation function: LeakyReLU -0.5

The training process is shown exemplary for the MedMNIST dataset in Fig. S4. Here we observe overall good convergence behavior. For Dirty-MNIST specifically we additionally employ a convergence aiding training procedure, in which we continuously scale up stochasticity in the network over the first 200 epochs. As this procedure is a variation of noisy training (NT) widely used for deterministic neural networks on analog hardware, we denote it here as incremental NT. For the Dirty-MNIST task a similar Hyperparameter search was conducted, with the following results, notably we utilize the schedulefree [55] version of RAdam[56]:

- Training Dataset: MNIST
- Data augmentation factor: 0
- train_batch_size: 256
- num_epochs: 2500
- num_train_samples: 4
- Loss function: TraceMeanField_ELBO
- learning rate: 0.00270
- weight_decay: 0.
- Optimizer: RAdamScheduleFree
- KL annealing schedule: linear
- KL max: 0.000211
- Activation function: LeakyReLU -0.5
- NT num_epochs: 200
- NT schedule: linear then plateau

Similarly to Fig. S4, the training process for Dirty-MNIST is shown in Fig. S5. We note that while the training shows similarly good convergence behavior to the MedMNIST results, we find differences in other aspects of the training: The learning curves are overall smoother, since the chosen optimizer reacts better to the stochastic environment of SVI in general. We further observe, particularly for the AUROC, that the training process is highly dynamic, which is mostly, but not completely addressed by KL annealing.

## Comparison between Hardware and Software Results

To compare how well the software model matches the photonic Bayesian machine we investigate how the key metrics for MedMNIST change over multiple samples for both the hardware and software setting. These findings are shown in Fig. S6. We observe that given enough samples the software model matches the photonic hardware extremely well. While both models show similar trends with the number of samples drawn, it appears that a minimum of ten samples is required to adequately capture the stochasticity of both SVI software model and the SVI hardware.

## Subsampling Dirty-MNIST

For Dirty-MNIST we only use 1/5th of full test dataset during prediction and equally sample randomly from all three test datasets (MNIST, Ambiguous-MNIST and Fashion-MNIST). To verify that even with the reduced dataset size we achieve comparable performance to the full datasets, we perform an ablation study on the software model. We assume this to be a good proxy for overall performance, due to the previously shown very good matching between software and hardware at 10 samples. The measured results are shown in Tab. S1. From this we conclude that subsampling has no significant impact on the final hardware results.

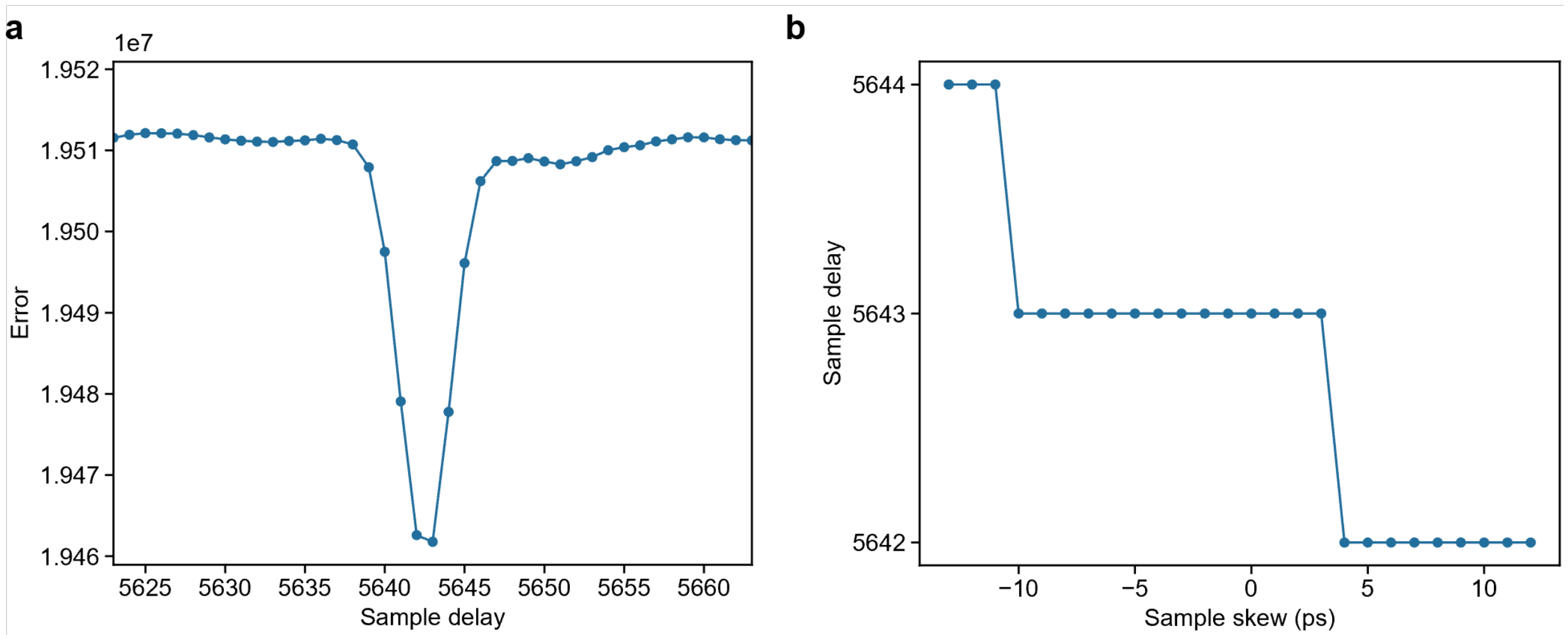

**Fig. S1: Delay and skew calibration.** **a**, We calibrate the sample delay between the triggering of the oscilloscope and the received data. There are two similarly good sample points as we use three samples to encode one symbol. **b**, We sweep the skew between the triggering of the oscilloscope and the capturing of the ADC to compensate for sub-sample delay. The optimal sample delay has a plateau-like shape given by the length of one sample of 12.5 ps. We choose the skew such that we are in the middle of a plateau.

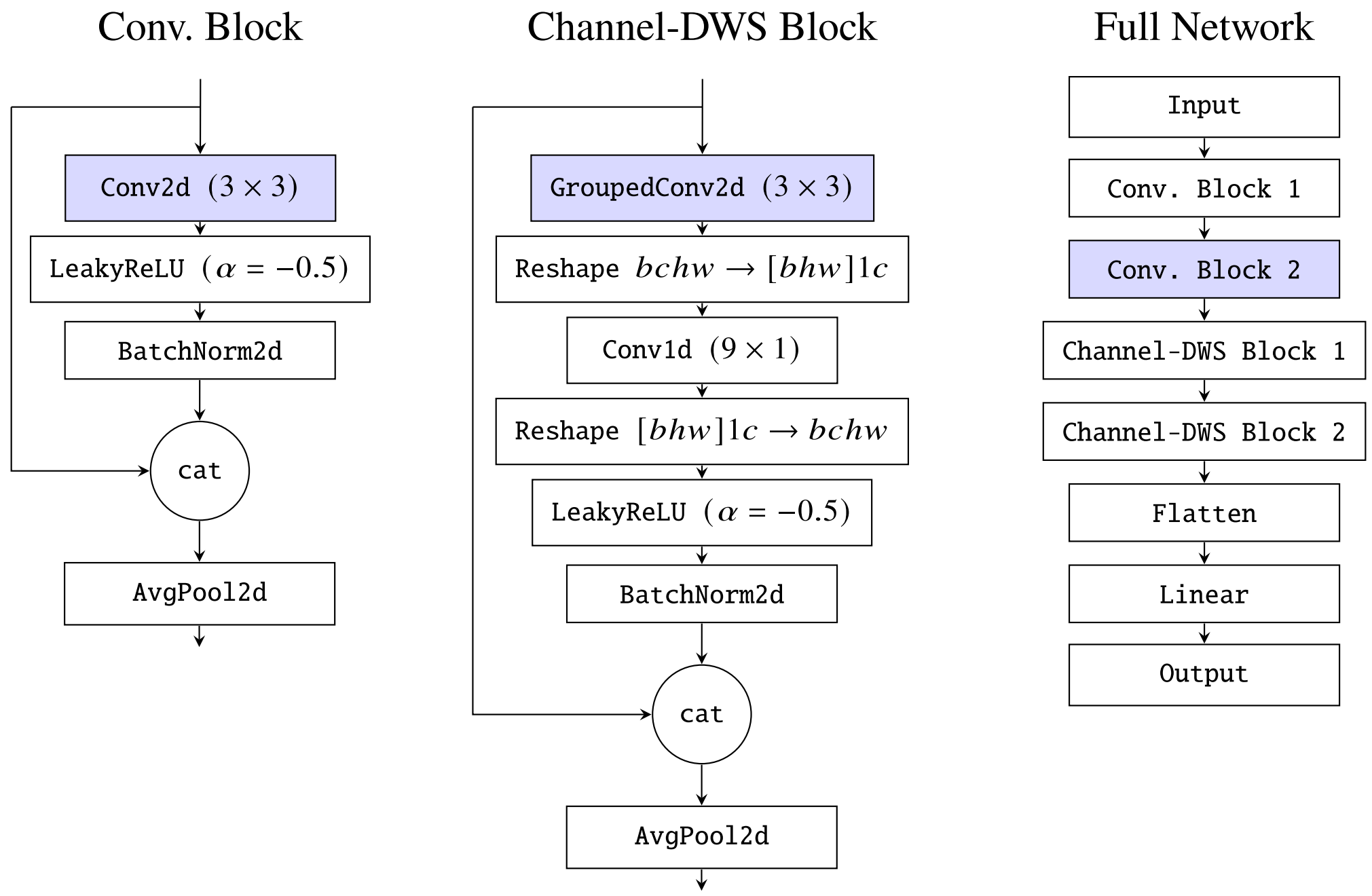

**Fig. S2: Hybrid photonic-digital Bayesian neural network.** We design a hybrid convolutional neural network for image classification, computing the deterministic layers digitally and offloading the probabilistic convolutions, marked in blue, to the photonic accelerator via its PyTorch interface. Two different types of convolutional blocks with skip connections are used to stack in total six convolutional layers, followed by a final linear layer. The size of the final linear layer depends on the number of different classes within the training dataset. All skip connections use concatenation over the channel dimension, as described by [18]. The DWS convolution operates as two convolutions, where the first is fully grouped, while the second one is a 1D Convolution over the channel dimension. Thus, we reorder the tensor shape of *Batch (b)*, *Channel (c)*, *Height (h)*, *Width (w)* to move the channels into the last dimension, while simultaneously moving all other parts into the first dimension and inserting a place holder dimension, denoted by *1*.

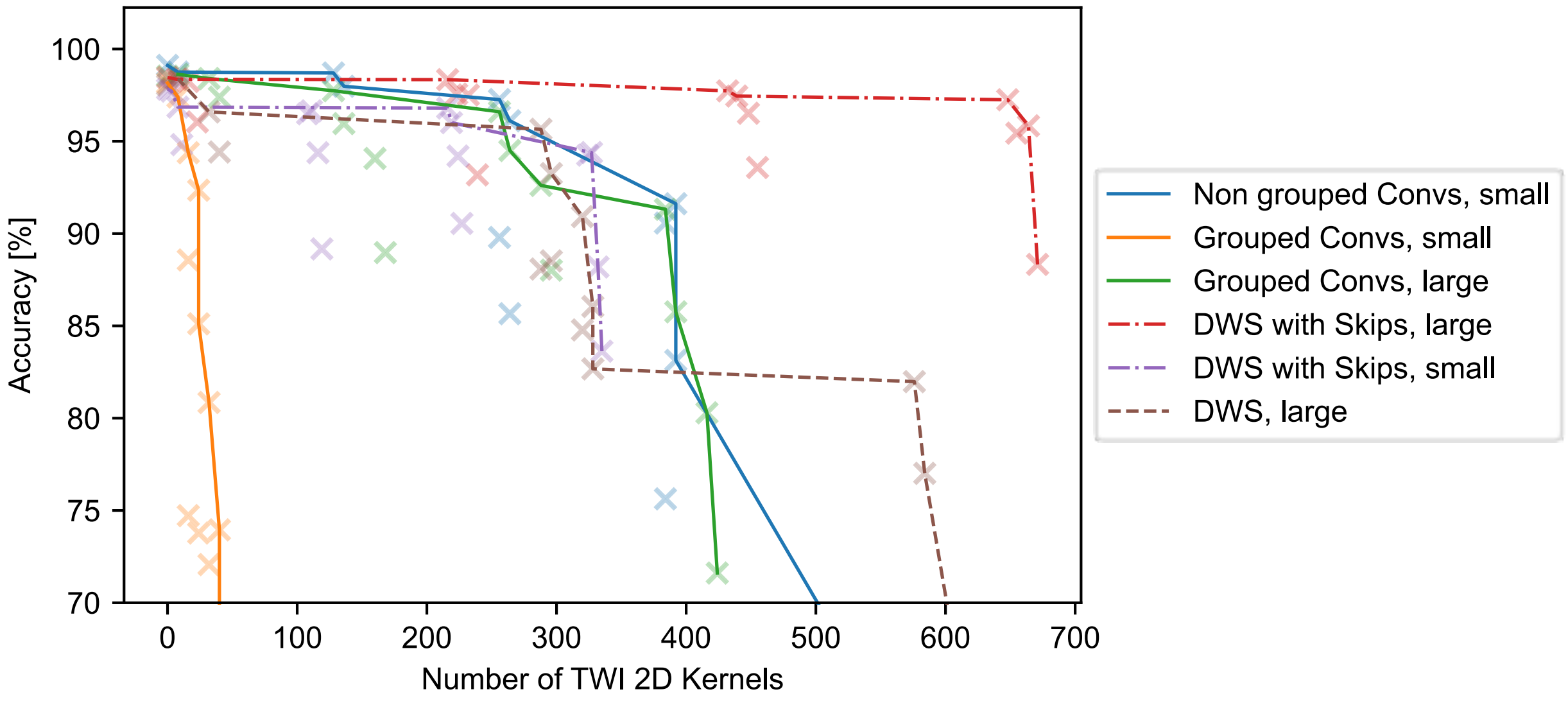

**Fig. S3: Architecture Search.** For multiple architecture candidates we plot their final test accuracy on MNIST against the number of photonic kernels in the network, when enabling different combinations of layers as probabilistic. Ideally, we try to maximize for both at the same time. For each candidate architecture we then draw a pareto frontier, making comparisons possible. Multiple conclusions can be drawn here: Grouped convolutions require more kernels compared to full convolutions to achieve the same accuracy; Depthwise Seperable (DWS) convolutions on their own make little difference to just grouped convolutions, however combining both DWS convolutions and skip connections enables smaller networks at the same accuracy, while scaling even better to larger networks. From these results we then choose the architecture with dominant pareto frontier for the final BNN architecture.

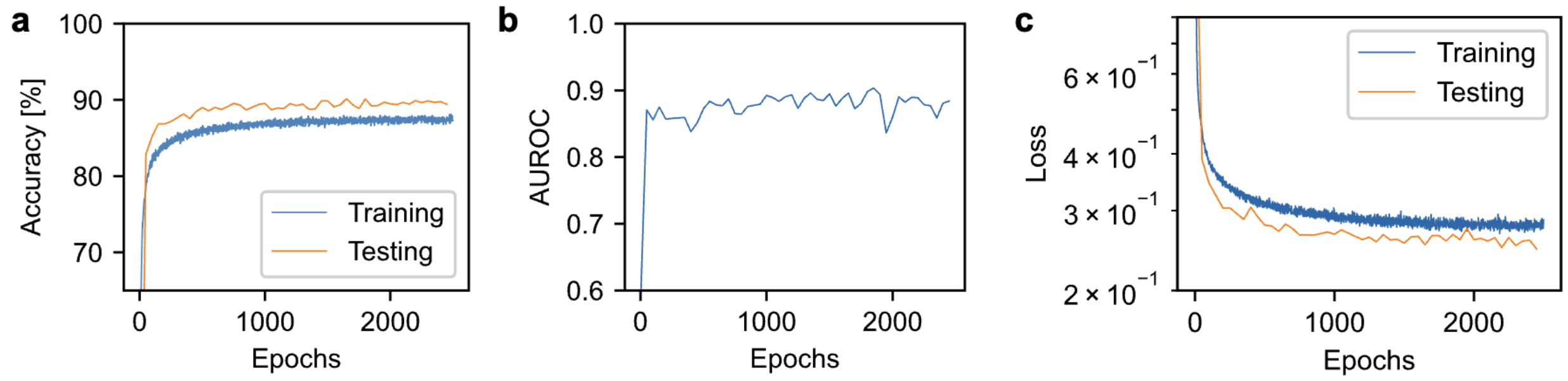

**Fig. S4: Bayesian Neural Network Training Process for MedMNIST. a**, The accuracy shows good convergence to an expected maximum. **b**, Similarly to the accuracy, the AUROC values against OOD data converge towards an optimal point over training. **c**, The loss curve during training shows very good convergence. We observe strong monotonicity and very little overfitting.

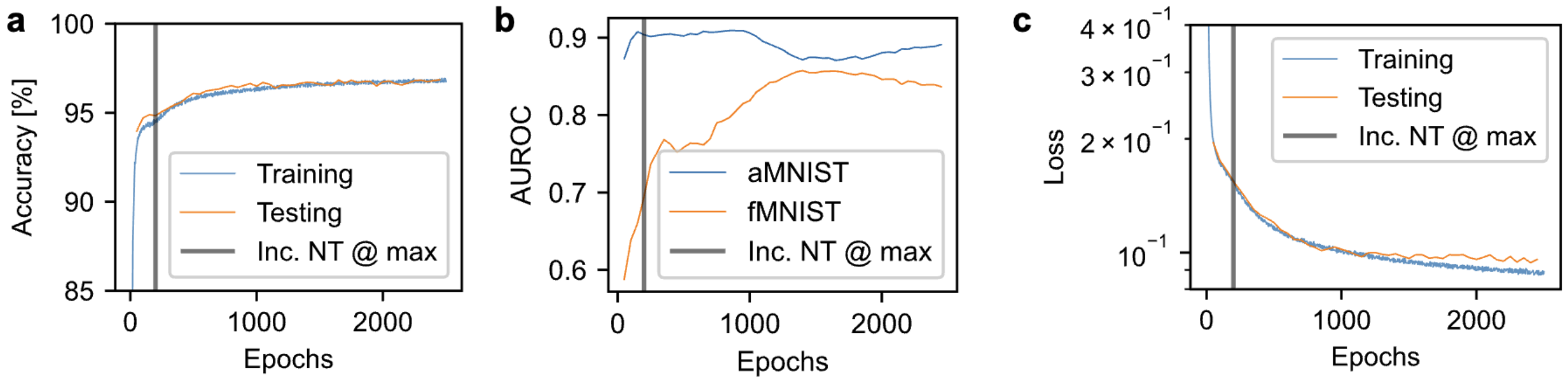

**Fig. S5: Bayesian Neural Network Training Process for Dirty-MNIST. a–c** In general, the training behaves like MedMNIST training in Fig. S4, however, we observe a significant impact of incremental noisy training (NT), where the end of the linearly increasing schedule is denoted as a black bar. Here we see that both the Accuracy and AUROC curves show significant changes and appear to benefit from an initial convergence to a high level. Interestingly the choice of optimizer results in an overall very smooth learning behavior, even under changing noise conditions and especially for the loss curve. Notably, we observe that the training process for Dirty-MNIST is much more dynamic for OOD detection than for MedMNIST. However, we still observe good convergence.

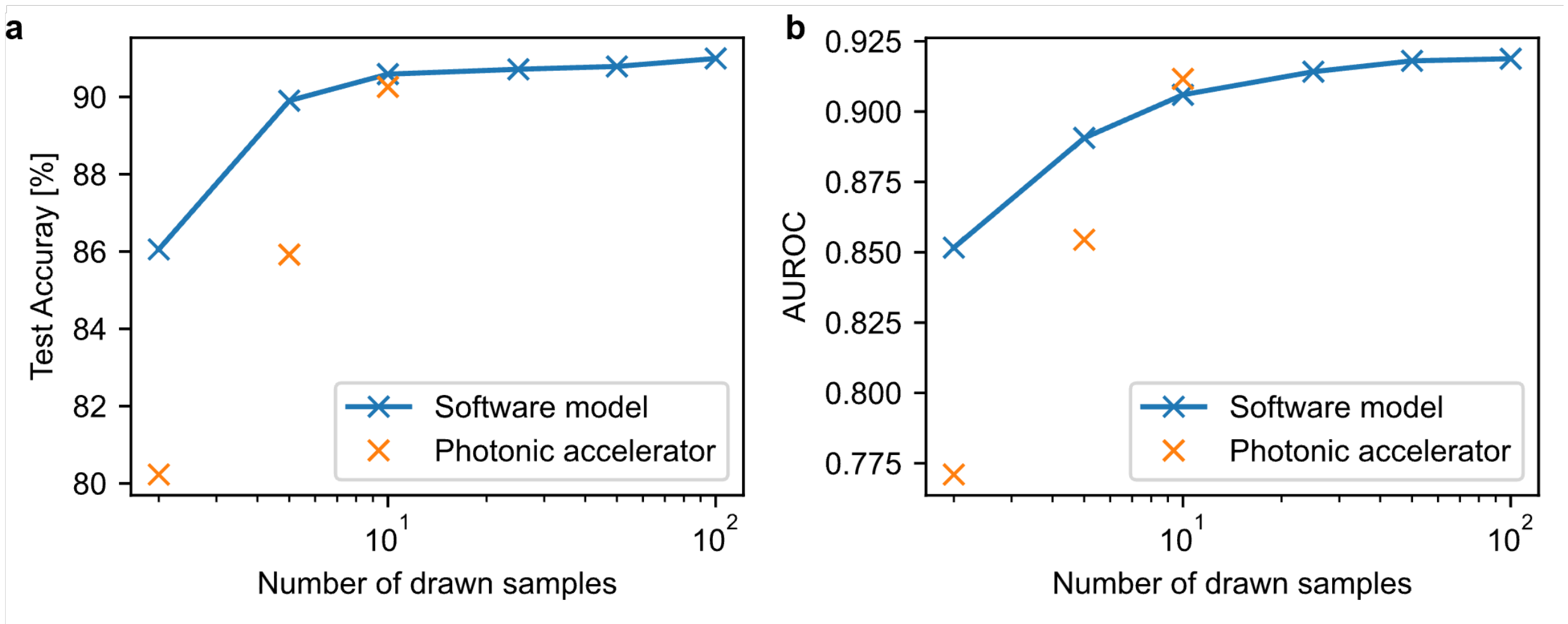

**Fig. S6: Comparison between the software model and photonic accelerator. a,b**, In the software model, the MedMNIST classification accuracy (**a**) and AUROC (**b**) saturate after approximately ten predictive samples. We therefore use ten predictive samples for experimental inference, as additional samples provide only marginal improvements. At this sample count, the hardware results agree closely with the software predictions.

**Table S1: Comparison of Full and Partial Dataset Inference.** The software model is only slightly disturbed by running on only 1/5th of the data. In fact, some metrics remain unaffected. From the close match between software and hardware we infer that the subsampling has negligible effect on the final hardware results.

| **Model** | **ID Accuracy [%]** | **AUROC f-MNIST [%]** | **AUROC a-MNIST [%]** |
|---|---|---|---|
| Photonic Accelerator | 96.01 | 84.4 | 88.0 |
| Software (1/5 data) | 96.91 | 85.6 | 89.6 |
| Software (full data) | 97.01 | 85.6 | 89.3 |